\documentclass[11pt]{article}

\usepackage[margin=1in]{geometry}
\usepackage[T1]{fontenc}
\usepackage[utf8]{inputenc}
\usepackage{amsmath,amssymb}
\usepackage{bm}
\usepackage{graphicx}
\usepackage{float}   
\usepackage{enumitem}
\usepackage[most]{tcolorbox}

\usepackage{booktabs}
\usepackage{caption}
\usepackage{microtype}
\usepackage{fancyvrb}
\usepackage{fvextra}   
\usepackage{parskip}   
\usepackage[numbers,sort&compress]{natbib}
\usepackage{url}
\usepackage[section]{placeins}   
\usepackage[hidelinks]{hyperref}

\graphicspath{{figures/}}

\definecolor{promptframe}{HTML}{8A8250}
\definecolor{promptback}{HTML}{FCFBEF}
\newtcolorbox{promptbox}[1]{%
  enhanced, breakable, colback=promptback, colframe=promptframe,
  boxrule=0.9pt, arc=3pt, left=7pt, right=7pt, top=6pt, bottom=6pt,
  fonttitle=\bfseries\small, coltitle=black,
  attach boxed title to top left={xshift=8pt, yshift=-8pt},
  boxed title style={colback=promptback, colframe=promptframe, boxrule=0.6pt, arc=2pt},
  title={#1}, top=12pt, before skip=4pt, after skip=4pt, fontupper=\small}
\newcommand{\pheading}[1]{\par\smallskip\noindent\textbf{#1}\quad}
\DefineVerbatimEnvironment{tcbverbatim}{Verbatim}%
  {fontsize=\scriptsize,xleftmargin=2pt,breaklines=true,breakanywhere=true,
   breaksymbolleft={},breaksymbolright={\tiny$\hookleftarrow$}}

\title{Can your AI agent be cheaper? Investigating the effects of task specifications on token spend in agentic coding tasks}
\author{Jakub Smékal \\ Stanford University}
\date{}

\begin{document}
\maketitle

\begin{abstract}
\noindent Agentic coding workflows are now widely deployed in real-world systems. With long-horizon reasoning and tool use, token usage has become an important consideration for both cost and efficiency. Two engineers using AI will solve the same problem differently. How the specification of a task shapes an agent's token spend, and whether that spend can be predicted in advance, are open questions. Here, we study the effects of different task specifications on agentic token spend with the Kimi K3 model at three thinking efforts. Across $2,700$ runs, we show that reducing a full task specification to a bare user story raises token spend by $29.7\%$, while run-to-run variance remains unaffected by any prompt changes. We show that prompt-sensitivity is task-dependent, running from $13\%$ to $115\%$. We fit a simple predictor that can price a full distribution of task specifications and thinking effort configurations from a single cheap probe on an unseen task within $36\%$, improving over prior work in predicting token spend. Our work provides initial results quantifying the effects of task specification on agentic token spend and introduces a method that can be used to systematically evaluate the cost of AI coding workflows.

\end{abstract}

\section{Introduction}

Token spend has become a central consideration in deploying agentic coding systems. Rising model capability unlocks longer-horizon autonomy, in which an agent works from a single task description without further human supervision. Recent work shows that agentic token usage is both large relative to single-shot use and inherently stochastic: two identical runs of the same prompt can differ by a factor of $30$ in tokens \cite{agents_money}.

Studies of agentic spend have varied the model, holding each task's problem statement fixed \cite{agents_money}. A separate line of work has varied the prompt, but only its surface form, holding its meaning constant \cite{prompteval,semantic_mutation}. The effect of a specific task description on token spend remains largely unmeasured. This poses two questions for practitioners:

\begin{itemize}
\item To what extent is token spend controllable through the task specification and the thinking effort setting?
\item To what extent is token spend predictable on a previously unseen task?
\end{itemize}

We hold the model fixed and vary the prompt. We construct a distribution of task specifications for a set of agentic coding tasks from SWE-bench Verified~\cite{swebench,swebench_verified}, bounded by an oracle specification that hands the agent the fix and a raw prompt of unstructured failing-test output, with structured variants in between that strip either most of the specification or one section at a time. Every specification is run at three thinking efforts, and each is repeated multiple times to estimate run-to-run token spread. Our results are summarized as follows:

\begin{itemize}
\item \textbf{The prompt moves the mean.} Cutting a full specification down to a plain prose description of the problem raises token spend by $29.7\%$ and turns to success by $16.4\%$, in the same direction on every task we measure.
\item \textbf{The prompt does not move the variance.} Rerunning an identical specification produces $\times1.34$ spread in token spend, and no specification we tested widens or narrows it.
\item \textbf{The cost distribution is cheaply predictable.} A single probe run costing eleven cents on an unseen task predicts its spend across every other specification and thinking effort to typically $36\%$, against $161\%$ with no measurement at all.
\end{itemize}

The rest of the paper describes our experimental methodology and results in greater detail. Section \ref{methods} describes how we construct a set of task specifications and our experimental setup, and Section \ref{results} presents our findings analyzing and predicting agentic token spend. We conclude with a discussion of takeaways for practitioners deploying agentic coding workflows.
\section{Related work}

\textbf{Cost as an outcome.} Inference cost is increasingly reported as a result rather than an implementation detail \cite{agents_matter,agentless}. Closest to our work, \cite{agents_money} studies eight frontier models on SWE-bench Verified and shows that runs on the same task differ by up to $30\times$ in tokens, that higher spend does not reliably buy accuracy, and that models estimate their own consumption poorly. That study varies the model while holding each task's problem statement fixed; we invert the design.

\textbf{The prompt as a variable.} Single-prompt benchmarks are fragile: performance swings by several points across paraphrases of one task \cite{prompteval,semantic_mutation}. Prior work held task-relevant information fixed and varied surface form, without repeated sampling and without a cost outcome. We vary task detail deliberately and price each increment. The distinction matters because task-relevant information is not monotonically beneficial: removing detail can improve correctness by disrupting misleading lexical cues \cite{underspec_helps}, and packaged skill documents may raise token spend without improving performance \cite{swe_skills}.

\textbf{What a task description should contain.} Bettenburg et al. \cite{bettenburg2008} showed that the most valued elements of a bug report are also the hardest to supply. Khatib et al. \cite{khatib2026} transpose this to agents over 433 SWE-bench Verified issues, finding that fix suggestions, reproduction scripts and localization are associated with higher resolution odds. \emph{SWE-Bench Pro} \cite{swebench_pro} pairs each task with human-written requirements. Related work lets the agent request additional information rather than varying the input prompt \cite{ambig_swe,clareval,value_of_information}. Here, we address the complementary question of how variations in the input task specification affect token spend.

\textbf{Thinking effort.} A parallel literature treats reasoning budget as the quantity to optimize  \cite{reasoning_budget}, either benchmarking allocation \cite{optimal_thinking} or learning it \cite{adaptive_ttc,selfbudgeter}. All of it takes the query as given, which our results suggest is incomplete, since how much a thinking budget buys depends on how the task was described.


To our knowledge, no prior work varies the level of detail in a task description while measuring token cost and its run-to-run variance.

\section{Methods}\label{methods}

\subsection{Tasks and prompts}

We take five tasks from SWE-bench Verified~\cite{swebench,swebench_verified}. Where prior benchmarking describes each task with a single prompt, we use fewer tasks but construct a distribution of prompts spanning different levels of task-relevant information, which aims to more closely mimic the variety of task specifications in real-world workflows and to let us attribute changes in token spend to the removal of specific task sections.

The specification set comprises ten spec variations and two anchor prompts. Each spec variation is constructed from the task's original SWE-bench problem statement, with structure derived from the GitHub Spec Kit template \cite{speckit}. A full specification has eight sections: header, user story, acceptance scenarios as Given/When/Then cases, edge cases, functional requirements, key entities, success criteria, and assumptions. The ten variations are the full specification, seven removing one section each, and two partial specifications retaining the header plus the user story, or the header plus requirements and success criteria. One variation removes the User Scenarios and Testing block entirely, meaning both the user story and its scenarios, since the scenarios are written in terms of the story; another removes the scenarios but keeps the story prose. The difference between these two variants is the Given/When/Then cases alone. Appendix~\ref{app:B} gives the ablation matrix and the length of every specification.


\begin{figure}[!ht]
\centering

\begin{promptbox}{\texttt{minimal} \textnormal{— header and user story only, 129 tokens}}
\pheading{Feature Specification} Preserve coordinate dtype through \texttt{stack}\\
\textbf{Feature Branch}: \texttt{fix/stack-preserves-coord-dtype} \quad
\textbf{Created}: 2022-12-20 \quad \textbf{Status}: Draft

\pheading{User Scenarios \& Testing}
\emph{User Story 1 — Stack without changing coordinate dtypes (Priority: P1)}

As a user stacking dimensions into a MultiIndex, I need the coordinate dtypes to
survive the operation, so that comparisons and downstream code that depend on a
narrow integer type keep working.

\textbf{Why this priority}: The change is silent; no error is raised, values
simply come back with a wider dtype.

\textbf{Independent Test}: Build a dataset with an \texttt{int32} coordinate,
stack it, and compare the coordinate dtype before and after.
\end{promptbox}

\vspace{0.6em}

\begin{promptbox}{\texttt{contract} \textnormal{— header, requirements and success criteria, 166 tokens}}
\pheading{Feature Specification} Preserve coordinate dtype through \texttt{stack}\\
\textbf{Feature Branch}: \texttt{fix/stack-preserves-coord-dtype} \quad
\textbf{Created}: 2022-12-20 \quad \textbf{Status}: Draft

\pheading{Functional Requirements}
\begin{itemize}[leftmargin=1.2em,itemsep=1pt,topsep=2pt]
  \item \textbf{FR-001}: The system MUST preserve each coordinate's dtype when a
        MultiIndex is created by \texttt{stack}.
  \item \textbf{FR-002}: The system MUST NOT alter values, only guarantee the
        dtype is carried through.
\end{itemize}

\pheading{Success Criteria}
\begin{itemize}[leftmargin=1.2em,itemsep=1pt,topsep=2pt]
  \item \textbf{SC-001}: A coordinate created with dtype \texttt{i4} still
        reports dtype \texttt{i4} after the dataset is stacked.
  \item \textbf{SC-002}: No existing stack or unstack behavior regresses.
\end{itemize}
\end{promptbox}

\caption{Two of the ten specification variants for \texttt{pydata/xarray-7393}. \texttt{minimal} retains only the narrative; \texttt{contract}
retains only the requirements and the criteria for meeting them. They describe
the same defect and are of comparable length. The remaining variations are shown in Figure~\ref{fig:B1}.}
\label{fig:1}
\end{figure}

The two anchors are constructed to bound the specification set.
The raw anchor provides contrast on structure; it is the failing test output, the least structured description that still converges on the correct fix. It is a realistic specification, since it is what an engineer pasting a raw error log would supply, and we include it in the analyses. The oracle anchor bounds task-relevant information; it states the solution and asks only that it be applied. We report the oracle as a sanity check that it is indeed the cheapest prompt, at $0.05$ to $0.19$ times the cost of the full specification, but exclude it from the analyses, as a real specification is unlikely to contain the solution, and including a variant that cheap by construction would inflate every effect we report. Analyses below therefore use eleven specifications. The exception is the token accounting in Figure~\ref{fig:5}, which covers every run.

Specifications were drafted with a language model, Fable 5, from the original task descriptions, then hand-edited for consistency and faithfulness to the template.

\subsection{Grid and execution}

Each task is paired with each of the twelve specifications at three thinking efforts (low, high, max), with fifteen repeats, giving $5 \times 12 \times 3 \times 15 = 2{,}700$ runs. All use Kimi K3~\cite{kimiteam2026kimik3openfrontier} through a Modal endpoint at temperature $1.0$. The scaffold is mini-swe-agent, running each task in the standard SWE-bench Docker image. Runs execute without network access, and every task is screened for solution leakage.

\subsection{Outcomes and analysis}

We record cost in USD at list prices, input, cached and output tokens, agent turns, and whether the patch resolves the task under the SWE-bench harness. Cost and tokens are near-equivalent on one price schedule, so we report cost and give token figures in Appendix~\ref{app:F}. 

For each section and outcome we fit one Bayesian hierarchical model, estimating a typical effect across tasks together with how far the tasks disagree about it, and report a posterior median and 90\% credible interval. Appendix~\ref{app:D} gives the model, its priors, and a sensitivity analysis.

\subsection{Prediction model}

Holding out one task at a time, let $y_{tbe}$ be the mean log cost of task $t$ under specification $b$ at effort $e$, and $\bar{y}_t$ its mean over all cells. From the four training tasks $T'$ we learn a shared shape

\begin{equation}
\hat{s}_{be} = \frac{1}{|T'|}\sum_{t \in T'} \left( y_{tbe} - \bar{y}_{t} \right),
\end{equation}

which records which configurations are relatively expensive while carrying no information about any task's overall level. For held-out task $h$ we take $k$ probe runs at a fixed configuration $(b_0, e_0)$ with log costs $z_1, \dots, z_k$, and estimate that level as

\begin{equation}
\hat{\delta} = \frac{1}{k}\sum_{i=1}^{k} z_i - \hat{s}_{b_0 e_0}, \qquad \hat{y}_{hbe} = \hat{\delta} + \hat{s}_{be},
\end{equation}

setting $\hat{\delta} = 0$ when $k = 0$. We configured the probe with a full task specification run at low thinking effort.

\section{Results}\label{results}

\subsection{Prompt content and token spend}

Prompt variation moves average token spend, in two forms that are near-interchangeable: across the $2{,}390$ solved runs of the eleven specification variations, log turns and log cost correlate at $r = 0.953$, so turns serve as a unit-free proxy for spend. Here, we report both. Every effect below carries a 90\% credible interval from a single hierarchical model (Appendix~\ref{app:D}).

\begin{figure}[!ht]
\centering
\includegraphics[width=0.92\linewidth]{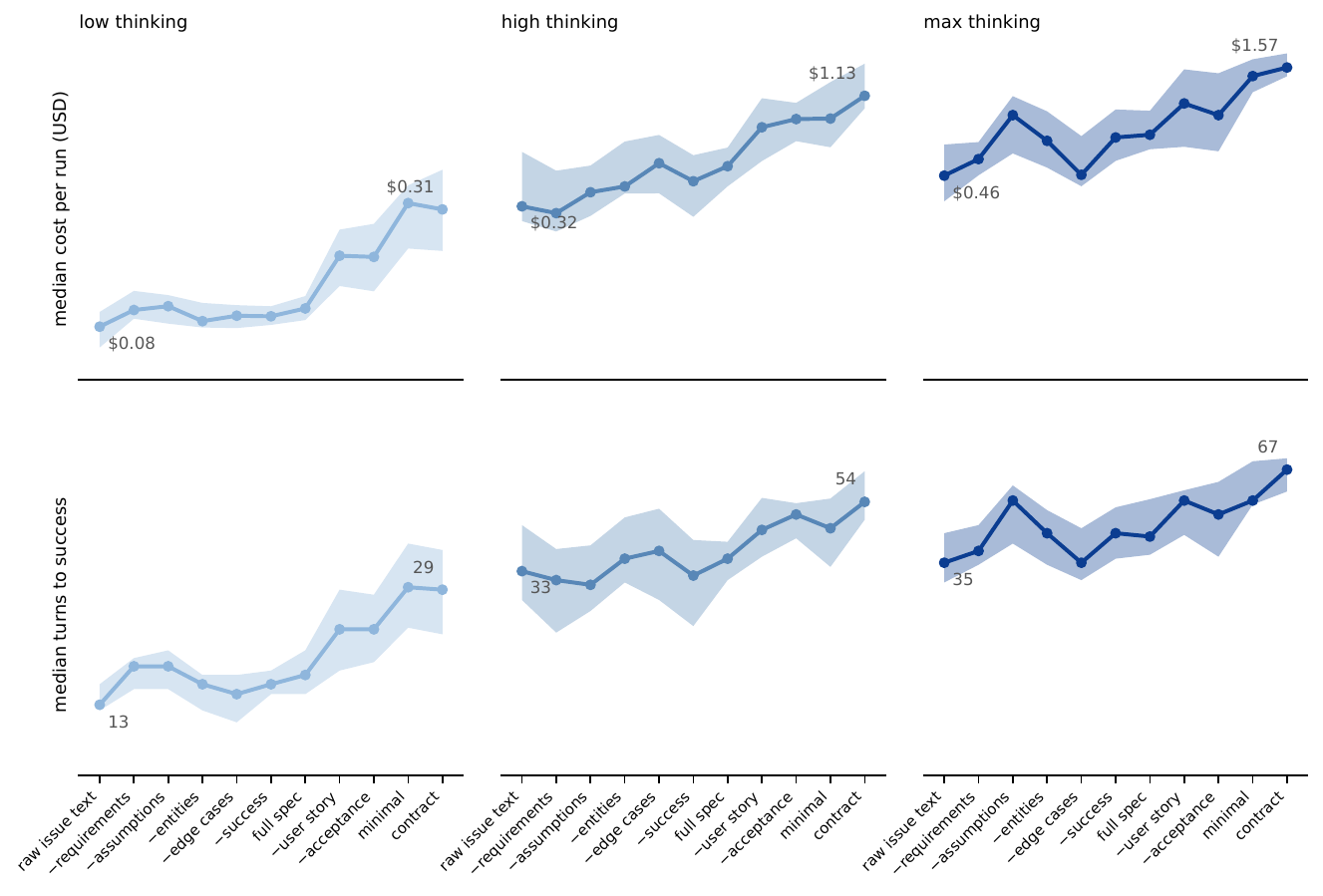}
\caption{Median cost per run (top) and turns to success (bottom) for xarray-7393, across eleven specification variants at three thinking efforts. Bands are interquartile ranges over 15 repeats.}
\label{fig:2}
\end{figure}

Per-section differences are small. Removing any one of functional requirements, assumptions, edge cases, key entities or success criteria changes cost by between $-5.4\%$ and $-2.4\%$ and turns by between $-1.6\%$ and $-0.4\%$. Larger ablations have a greater effect. Reducing the specification to a bare user story raises cost by $29.7\%$ and turns by $16.4\%$, positive on all five tasks. On xarray-7393 the same ablation raises cost by $115\%$, four times the pooled effect; that task is an outlier, but it establishes that tasks exist on which the prompt has considerable leverage, which is why a practitioner should measure their own task rather than assume its cost sensitivity a priori. The only single section whose removal consistently has an isolated effect is the acceptance scenarios, at $7.0\%$ additional turns. Between-task disagreement rises with the size of the ablation: $4\%$ to $8\%$ for the five minor single-section removals, $22\%$ and $26\%$ for acceptance scenarios and user story-with-scenarios, and $29\%$ and $40\%$ for the two largest cuts (Appendix~\ref{app:E}). Notably, user story-with-scenarios shows this elevated disagreement despite having no established average effect on any outcome.

\begin{figure}[!ht]
\centering
\includegraphics[width=1.0\linewidth]{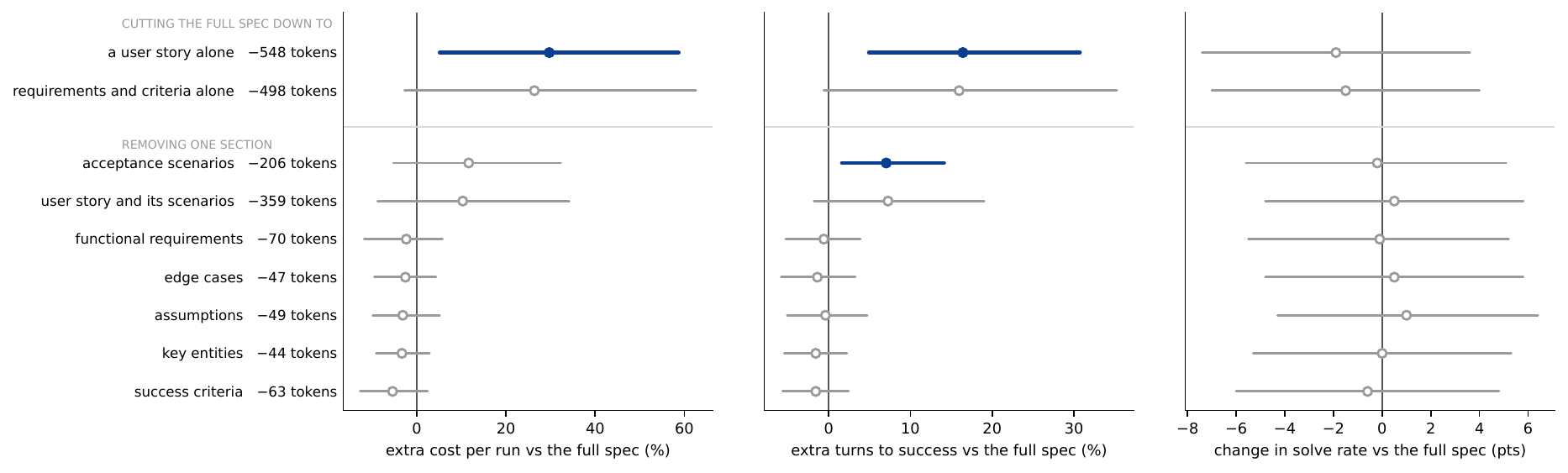}
\caption{Effect of cutting the full specification, pooled across five tasks and three thinking efforts. The upper pair cuts many sections from the full specification; the rest remove a single section at a time. Each label provides the number of tokens removed in the variation, averaged over the five tasks. Dots are posterior medians, lines are 90\% credible intervals, and blue marks intervals excluding zero. No section measurably changes solve rate, every interval lying within 7.5 points of zero, so differences in cost and turns are attributable to the prompt rather than to some variants solving an easier version of the task.}
\label{fig:3}

\end{figure}

Acceptance scenarios and success criteria state the same requirement at different levels of abstraction: the first as executable Given/When/Then cases, the second as prose assertions about the same behavior. Removing the concrete form costs turns on all five tasks; removing the abstract form has no measurable effect on any outcome. What matters is concreteness about which cases must pass, not the presence of a stated requirement. The raw anchor supports this interpretation. The least structured prompt is the cheapest of the eleven variations at low effort on four of five tasks, at $0.73\times$ to $1.10\times$ the full specification pooled over efforts. A failing-test transcript names the file and test that must pass, and that localization substitutes for the discovery turns a prose specification leaves to the agent.

\subsection{Thinking effort}

Prompt variation matters more at low thinking effort than at max. Pooled over tasks, the ratio between the most and least expensive specification narrows from $\times2.13$ at low effort to $\times1.67$ at high and $\times1.61$ at max, and the cost of removing the acceptance scenarios falls from $20.1\%$ additional turns to $4.5\%$, and then $2.1\%$, respectively. The pattern holds on four of the five tasks; on astropy-14365 the ratio only moves from $\times1.57$ at low to $\times1.47$ at max effort.

The reading is that thinking effort and specification detail are substitutes only where thinking is scarce. At low effort, information withheld from the prompt is recovered by model reasoning, and that reasoning is what the missing section costs. At max effort the model reasons extensively regardless of the prompt details, so supplying the same information changes little. Geometric mean spend rises from $0.117$ USD per run at low effort to $0.561$ USD at max, so the gap in USD between cheapest and most expensive specification widens even as the relative ratio falls.

Specification had no credible effect on solve rate: the $90\%$ credible interval for every section removed includes zero (Table~\ref{tab:E1}), unlike cost and turns, where the two largest cuts are credibly positive. Four of the five tasks solve at $98\%$ or above; django-15503, at $86.3\%$, is the only task below that range. Per-task and per-effort solve rates are given in Table~\ref{tab:A1} and Figure~\ref{fig:F3}.

\subsection{Run-to-run spread}

\begin{figure}[!ht]
\centering
\includegraphics[width=1.0\linewidth]{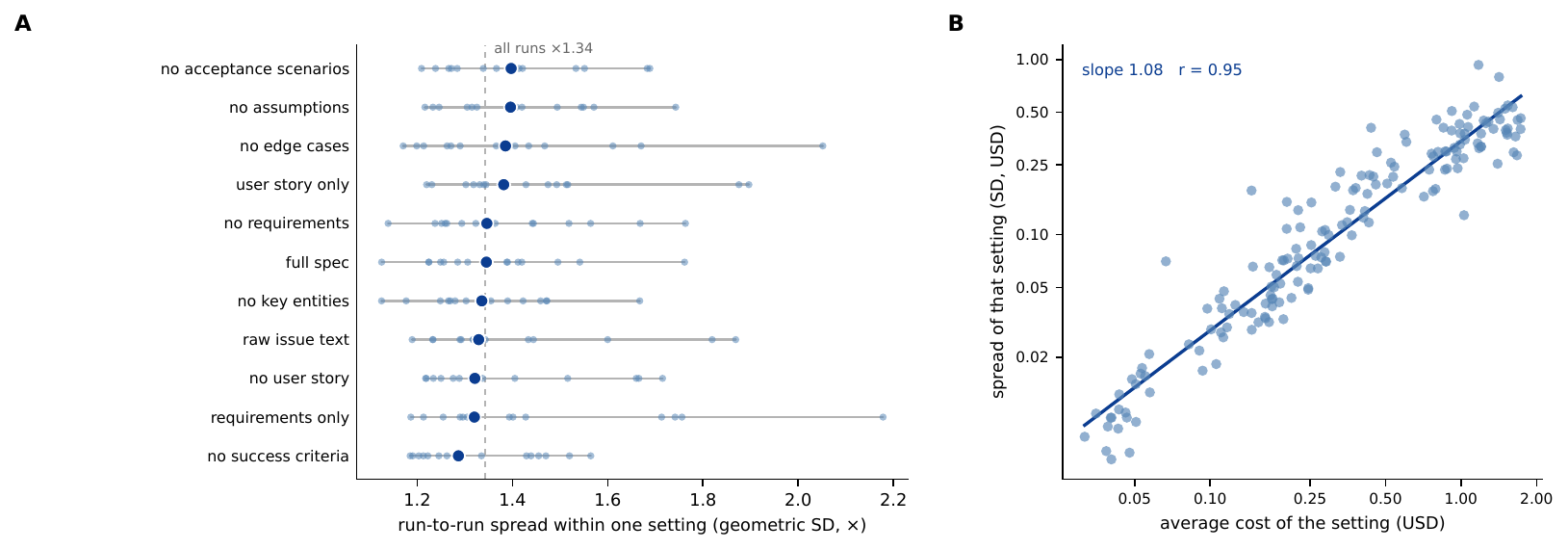}
\caption{A: run-to-run spread within a single setting, one dot per task and thinking effort; the dark marker is each specification's median and the dashed line the median across all settings, \emph{$\times1.34$}. B: spread of a setting in dollars against its average cost. The slope of \emph{$1.08$} is close to proportionality, so how much a run varies in absolute terms is set by its mean cost, not by the task description.}
\label{fig:4}
\end{figure}

 Prompt variation had no measurable effect on variance. Within a single specification and effort, repeats have a median geometric standard deviation of $\times1.34$. Taking each specification separately, median geometric standard deviation ranges from $\times1.29$ to $\times1.40$, which is a small enough gap that the underlying per-task-and-effort values overlap heavily across specifications (Figure~\ref{fig:4}A). What variation exists is explained by cost level: absolute spread scales almost proportionally with average cost, at a log-log slope of $1.08$ and $r = 0.95$ (Figure~\ref{fig:4}B), so more expensive settings are inherently more stochastic in cost. A per-cell comparison against a null built from sampling noise alone found no reliable excess beyond what finite-sample estimation error predicts, consistent with cost-driven variance.

This is consistent with prior work that agentic spend is inherently stochastic; \citet{agents_money} find identical repeats differing by up to a factor of $30$ across all eight models they evaluate. The most reliable regimes we found were simply the cheapest, since absolute variance falls with the mean. This suggests that the only route to a more predictable agentic token spend is a smaller one.

\begin{figure}[!ht]
\centering
\includegraphics[width=0.80\linewidth]{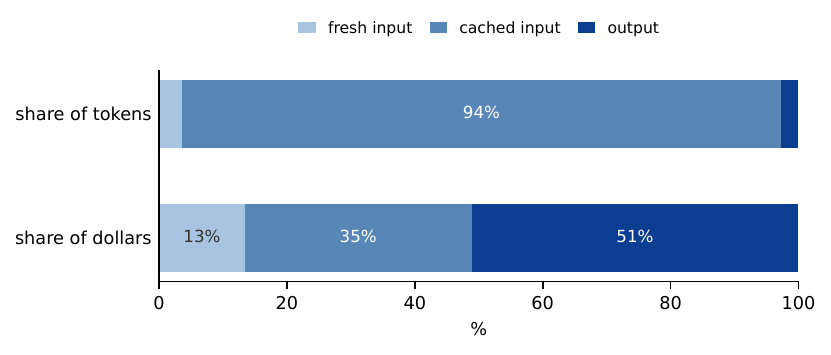}
\caption{Share of tokens against share of dollars, by token class, over all 2,700 runs at Kimi K3 list prices.}
\label{fig:5}
\end{figure}

One structural feature determines which cost optimizations can matter (Figure~\ref{fig:5}). In our experiments with Kimi K3, output tokens are $2.7\%$ of tokens processed but $51.1\%$ of dollars at a $96.3\%$ cache hit rate, and fresh input is $13.4\%$ of spend. Optimizations aimed at the input, such as prompt compression or context trimming, therefore address a small fraction of spend, whereas reducing turns addresses the majority of cost. The split depends on the price schedule and cache hit rate rather than on the agent alone, so these percentages do not directly apply to other models or inference stacks.

\subsection{Predicting the cost of an unseen task}

We learn a shared shape over the prompt $\times$ effort grid from four tasks, hold out the fifth, and use $k$ probe runs at one fixed setting to fix its level. Holding out one task at a time leaves 32 other settings per task to predict, from 11 specifications × 3 efforts minus the probe, or 160 held-out settings in total across the five folds. Each figure below is a median over those 160 settings, averaged over 400 independent probe draws. Median error is the typical gap between predicted and actual cost; the budget multiplier is the factor by which a prediction must be inflated for the true cost to fall under it in $90\%$ of settings.

\begin{figure}[!ht]
\centering
\includegraphics[width=1.0\linewidth]{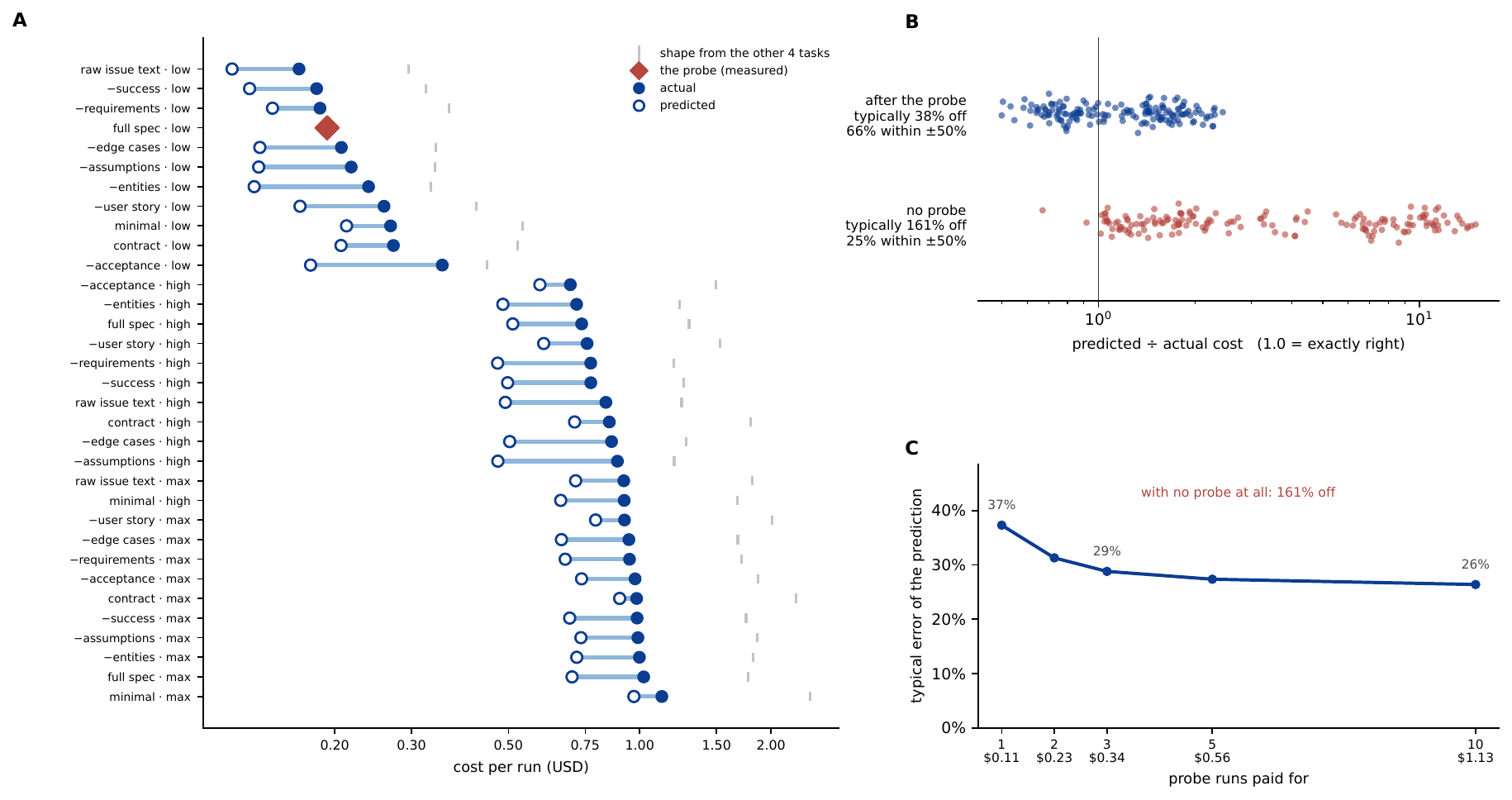}
\caption{Prediction of held-out task cost. A: worked example on astropy-13579, actual against predicted cost for every setting, with the probe marked. B: prediction error with and without the probe. C: error against the number of probe runs.}
\label{fig:6}
\end{figure}

\begin{table}[!ht]
\centering
\setlength{\tabcolsep}{4pt}
\small
\begin{tabular}{@{}lrrrr@{}}
\toprule
probe runs & probe cost (USD) & median error & settings within $\pm50\%$ & budget multiplier for $90\%$ \\
\midrule
0 & 0 & $161\%$ & $25\%$ &  \\
\textbf{1} & \textbf{0.11} & $36\%$ & $67\%$ & $1.9\times$ \\
3 & 0.34 & $29\%$ & $74\%$ & $1.8\times$ \\
10 & 1.13 & $25\%$ & $76\%$ & $1.7\times$ \\
\bottomrule
\end{tabular}
\caption{Prediction of held-out task cost against the number of probe runs.}
\label{tab:1}
\end{table}

The probe determines the level of a task rather than the relative cost of its configurations, which is fit from previous tasks. Without the probe, predicting a new task from the other four is a median of $161\%$ off; one run at $0.11$ USD cuts that to $36\%$ and brings two thirds of settings within $\pm50\%$, while ten probes reach $25\%$ median error, as shown in Table~\ref{tab:1}.

\citet{agents_money} asked the agent to estimate its own usage before executing the task and reported the correlation between predicted token spend and actual token spend. We now report our results using the same metric, where we once again see a gap between inferring the cost from previous tasks without measuring vs. calibrating the cost curve to an unseen task by running a measurement probe.
Scoring individual runs within a single thinking level, prediction without a probe reaches $r = 0.08$ ($0.04$ to $0.12$, bootstrapped over runs), at or below the $0.04$ to $0.39$ in prior work~\cite{agents_money}. One probe run takes the same predictor to $r = 0.72$ ($0.67$ to $0.78$ across draws of the probe run), improving over prior work, although it requires measuring the unseen task rather than asking a model to reason about its cost before running it.
\section{Discussion}

Mean token spend can be shifted by changes in task specification, but the size of that shift varies considerably from task to task: the same cut costs $13\%$ on one task and $115\%$ on another. Our study uses a single model, which is a core limitation. Where comparison is possible, however, our findings agree with studies that vary the model while fixing one prompt per task \cite{agents_money}: repeated runs of an identical input remain highly stochastic in token spend, and prior work established that greater spend does not reliably buy correctness.

For practitioners, our results support the utility of measuring against a prompt distribution to evaluate and predict token cost for a given task. A small set of tasks, sampled across specification variants and thinking efforts, gives the shape of the cost curve for a workload; one probe run then calibrates it to a new task. This method may help practitioners quantify which AI usage patterns are expensive and how much of that expense is recoverable by prompt choice.

Specification detail and thinking effort interact. Prompt leverage falls as effort rises, from $\times2.13$ to $\times1.61$, while the cost of removing the acceptance scenarios falls from $20.1\%$ additional turns to $2.1\%$. This bears on the practice of shipping large packaged instruction sets, such as skill documents and prompting frameworks, alongside every request: those are paid for on every run, whereas the work they save is realized only where the model would not otherwise have reasoned its way to the same outcome, and that margin narrows precisely in the high and max-effort configurations where such packages are most often deployed. This is consistent with reports that added skill documents raise token spend without improving performance \cite{swe_skills}.

Predicting a new task by inferring a cost shape from four others, without running it, is a median of $161\%$ off, and on the correlation metric used by prior work it is no better than a model's own estimate of its consumption \cite{agents_money}. A single run at $0.11$ USD reduces that to $36\%$. Cheap measurement can therefore be useful whenever the cost of a counterfactual matters, e.g., if a first prompt variation did not produce the desired result and the cost of another must be weighed, or when an agentic workflow must be judged against the rest of the distribution, as in repeatable asynchronous tasks running from the same base prompt. A single probe fixes the cost level for the other configurations of a task, so the measurement is made once per task rather than once per configuration.

Run-to-run spread appears to be a property of the serving stack and the model rather than the prompt. Repeats of an identical specification vary by $\times1.34$, no specification narrows that, and absolute spread scales almost exactly with mean cost, so a more predictable spend is obtained by making runs cheaper. At a $96.3\%$ cache hit rate, output tokens are $2.7\%$ of tokens processed and $51.1\%$ of money spent, so optimizations aimed at shortening the input address a minority of cost while reducing turns addresses the majority.

Our work considers a sparse set of agentic coding tasks while increasing the number of per-configuration repeats, more than 3× over prior work~\cite{agents_money}. Task sparsity and the use of a single model remain the key limitations on generalizing to a broader set of models and real-world tasks. Currently, our method for inferring a shared cost structure across prompt variants omits any effect of task-specific prompt sensitivity. We leave this to future work.

Our findings quantify a question that arises frequently in practice but is rarely measured: two engineers handed the same task will describe it differently and use different model configurations, both of which carry different costs. The method we introduce measures the spread in token spend across different descriptions of the same task, and we predict the cost of a new task from a single cheap probe run. We hope to extend our work to more tasks and models, as well as to real workflows, where task difficulty varies more widely and specifications can be collected from engineers rather than constructed.

\section{Conclusion}

With increasing adoption of agentic coding workflows, token usage becomes a primary consideration for many practitioners. Here, we evaluated token spend in agentic coding tasks across a set of different task specifications, and showed how a cheap probe can be used to predict the cost of a new task. Our results suggest that there are tasks for which token spend can be substantially improved through careful task specification. We hope our method motivates a more systematic evaluation of how usage patterns affect token spend in agentic coding workflows.

\bibliographystyle{plainnat}
\bibliography{references}

@article{agents_money,
  author       = {Bai, Longju and Huang, Zhemin and Wang, Xingyao and Sun, Jiao and
                  Mihalcea, Rada and Brynjolfsson, Erik and Pentland, Alex and Pei, Jiaxin},
  title        = {How Do {AI} Agents Spend Your Money? {A}nalyzing and Predicting Token
                  Consumption in Agentic Coding Tasks},
  journal      = {arXiv preprint arXiv:2604.22750},
  eprint       = {2604.22750},
  archivePrefix= {arXiv},
  primaryClass = {cs.SE},
  year         = {2026}
}

@article{agents_matter,
  author       = {Kapoor, Sayash and Stroebl, Benedikt and Siegel, Zachary S. and
                  Nadgir, Nitya and Narayanan, Arvind},
  title        = {{AI} Agents That Matter},
  journal      = {arXiv preprint arXiv:2407.01502},
  eprint       = {2407.01502},
  archivePrefix= {arXiv},
  primaryClass = {cs.LG},
  year         = {2024}
}

@article{agentless,
  author       = {Xia, Chunqiu Steven and Deng, Yinlin and Dunn, Soren and Zhang, Lingming},
  title        = {Agentless: Demystifying {LLM}-based Software Engineering Agents},
  journal      = {arXiv preprint arXiv:2407.01489},
  eprint       = {2407.01489},
  archivePrefix= {arXiv},
  primaryClass = {cs.SE},
  year         = {2024}
}

@inproceedings{prompteval,
  author       = {Polo, Felipe Maia and Xu, Ronald and Weber, Lucas and Silva, M{\'i}rian and
                  Bhardwaj, Onkar and Choshen, Leshem and de Oliveira, Allysson Flavio Melo and
                  Sun, Yuekai and Yurochkin, Mikhail},
  title        = {Efficient Multi-Prompt Evaluation of {LLM}s},
  booktitle    = {Advances in Neural Information Processing Systems (NeurIPS)},
  year         = {2024},
  eprint       = {2405.17202},
  archivePrefix= {arXiv}
}

@article{semantic_mutation,
  author       = {Pan, Zhiyuan and Hu, Xing and Xia, Xin and Yang, Xiaohu},
  title        = {Re-Evaluating Code {LLM} Benchmarks Under Semantic Mutation},
  journal      = {arXiv preprint arXiv:2506.17369},
  eprint       = {2506.17369},
  archivePrefix= {arXiv},
  primaryClass = {cs.SE},
  year         = {2025}
}

@article{underspec_helps,
  author       = {Akli, Amal and Papadakis, Mike and Cordy, Maxime and Le Traon, Yves},
  title        = {When Prompt Under-Specification Improves Code Correctness: An Exploratory
                  Study of Prompt Wording and Structure Effects on {LLM}-Based Code Generation},
  journal      = {arXiv preprint arXiv:2604.24712},
  eprint       = {2604.24712},
  archivePrefix= {arXiv},
  primaryClass = {cs.SE},
  year         = {2026}
}

@article{swe_skills,
  author       = {Han, Tingxu and Zhang, Yi and Song, Wei and Fang, Chunrong and
                  Chen, Zhenyu and Sun, Youcheng and Hu, Lijie},
  title        = {{SWE}-Skills-Bench: Do Agent Skills Actually Help in Real-World
                  Software Engineering?},
  journal      = {arXiv preprint arXiv:2603.15401},
  eprint       = {2603.15401},
  archivePrefix= {arXiv},
  primaryClass = {cs.SE},
  year         = {2026}
}

@inproceedings{bettenburg2008,
  author       = {Bettenburg, Nicolas and Just, Sascha and Schr{\"o}ter, Adrian and
                  Weiss, Cathrin and Premraj, Rahul and Zimmermann, Thomas},
  title        = {What Makes a Good Bug Report?},
  booktitle    = {Proceedings of the 16th ACM SIGSOFT International Symposium on
                  Foundations of Software Engineering (FSE)},
  pages        = {308--318},
  year         = {2008},
  publisher    = {ACM},
  doi          = {10.1145/1453101.1453146}
}

@article{khatib2026,
  author       = {Khatib, Lara and Mathews, Noble Saji and Nagappan, Meiyappan and
                  Nie, Pengyu and Zimmermann, Thomas},
  title        = {What Makes a Good Bug Report for an {AI} Agent?},
  journal      = {arXiv preprint arXiv:2607.07593},
  eprint       = {2607.07593},
  archivePrefix= {arXiv},
  primaryClass = {cs.SE},
  year         = {2026}
}

@article{swebench_pro,
  author       = {Deng, Xiang and Da, Jeff and Pan, Edwin and He, Yannis Yiming and
                  Ide, Charles and Garg, Kanak and Lauffer, Niklas and Park, Andrew and
                  Pasari, Nitin and Rane, Chetan and Sampath, Karmini and Krishnan, Maya and
                  Kundurthy, Srivatsa and Hendryx, Sean and Wang, Zifan and
                  Bharadwaj, Vijay and Holm, Jeff and Aluri, Raja and
                  Zhang, Chen Bo Calvin and Jacobson, Noah and Liu, Bing and Kenstler, Brad},
  title        = {{SWE}-Bench Pro: Can {AI} Agents Solve Long-Horizon Software
                  Engineering Tasks?},
  journal      = {arXiv preprint arXiv:2509.16941},
  eprint       = {2509.16941},
  archivePrefix= {arXiv},
  primaryClass = {cs.SE},
  year         = {2025}
}

@inproceedings{ambig_swe,
  author       = {Vijayvargiya, Sanidhya and Zhou, Xuhui and Yerukola, Akhila and
                  Sap, Maarten and Neubig, Graham},
  title        = {{Ambig-SWE}: Interactive Agents to Overcome Underspecificity in
                  Software Engineering},
  booktitle      = {International Conference on Learning Representations (ICLR)},
  eprint       = {2502.13069},
  archivePrefix= {arXiv},
  primaryClass = {cs.SE},
  year         = {2026},
}

@article{clareval,
  author       = {Li, Jialin and Wu, Yuan and Chang, Yi},
  title        = {{ClarEval}: A Benchmark for Evaluating Clarification Skills of Code
                  Agents under Ambiguous Instructions},
  journal      = {arXiv preprint arXiv:2603.00187},
  eprint       = {2603.00187},
  archivePrefix= {arXiv},
  primaryClass = {cs.SE},
  year         = {2026}
}

@article{value_of_information,
  author       = {Dong, Yijiang River and Hu, Tiancheng and Hui, Zheng and Zhang, Caiqi and
                  Vuli{\'c}, Ivan and Bobu, Andreea and Collier, Nigel},
  title        = {Value of Information: A Framework for Human-Agent Communication},
  journal      = {arXiv preprint arXiv:2601.06407},
  eprint       = {2601.06407},
  archivePrefix= {arXiv},
  primaryClass = {cs.CL},
  year         = {2026}
}

@article{optimal_thinking,
  author       = {Aggarwal, Pranjal and Kim, Seungone and Lanchantin, Jack and
                  Welleck, Sean and Weston, Jason and Kulikov, Ilia and Saha, Swarnadeep},
  title        = {{OptimalThinkingBench}: Evaluating Over and Underthinking in {LLM}s},
  journal      = {arXiv preprint arXiv:2508.13141},
  eprint       = {2508.13141},
  archivePrefix= {arXiv},
  primaryClass = {cs.CL},
  year         = {2025}
}

@article{adaptive_ttc,
  author       = {Zhai, Zhiyuan and Li, Bingcong and Xiao, Bingnan and Li, Ming and Wang, Xin},
  title        = {Adaptive Test-Time Compute Allocation for Reasoning {LLM}s via
                  Constrained Policy Optimization},
  journal      = {arXiv preprint arXiv:2604.14853},
  eprint       = {2604.14853},
  archivePrefix= {arXiv},
  primaryClass = {cs.LG},
  year         = {2026}
}

@article{selfbudgeter,
  author       = {Li, Zheng and Dong, Qingxiu and Ma, Jingyuan and Zhang, Di and
                  Jia, Kai and Sui, Zhifang},
  title        = {{SelfBudgeter}: Adaptive Token Allocation for Efficient {LLM} Reasoning},
  journal      = {arXiv preprint arXiv:2505.11274},
  eprint       = {2505.11274},
  archivePrefix= {arXiv},
  primaryClass = {cs.CL},
  year         = {2025}
}

@article{reasoning_budget,
  author       = {Alomrani, Mohammad Ali and Zhang, Yingxue and Li, Derek and Sun, Qianyi and
                  Pal, Soumyasundar and Zhang, Zhanguang and Hu, Yaochen and
                  Ajwani, Rohan Deepak and Valkanas, Antonios and Karimi, Raika and
                  Cheng, Peng and Wang, Yunzhou and Liao, Pengyi and Huang, Hanrui and
                  Wang, Bin and Hao, Jianye and Coates, Mark},
  title        = {Reasoning on a Budget: A Survey of Adaptive and Controllable Test-Time
                  Compute in {LLM}s},
  journal      = {arXiv preprint arXiv:2507.02076},
  eprint       = {2507.02076},
  archivePrefix= {arXiv},
  primaryClass = {cs.CL},
  year         = {2025}
}

@misc{speckit,
  author       = {{GitHub}},
  title        = {Spec Kit},
  howpublished = {\url{https://github.com/github/spec-kit}},
  year         = {2025},
  note         = {Software}
}

@inproceedings{mini_swe_agent,
  title={{SWE}-agent: Agent-Computer Interfaces Enable Automated Software Engineering},
  author={John Yang and Carlos E Jimenez and Alexander Wettig and Kilian Lieret and Shunyu Yao and Karthik R Narasimhan and Ofir Press},
  booktitle={The Thirty-eighth Annual Conference on Neural Information Processing Systems},
  year={2024},
  url={https://arxiv.org/abs/2405.15793}
}

@inproceedings{swebench,
      title={{SWE}-bench: Can Language Models Resolve Real-World {GitHub} Issues?}, 
      author={Carlos E. Jimenez and John Yang and Alexander Wettig and Shunyu Yao and Kexin Pei and Ofir Press and Karthik Narasimhan},
      booktitle     = {International Conference on Learning Representations (ICLR)},
      year={2024},
      eprint={2310.06770},
      archivePrefix={arXiv},
      primaryClass={cs.CL},
      url={https://arxiv.org/abs/2310.06770}, 
}

@misc{swebench_verified,
  author       = {Chowdhury, Neil and Aung, James and Chan, Jun Shern and Jaffe, Oliver and
                  Sherburn, Dane and Starace, Giulio and Mays, Evan and Dias, Rachel and
                  Aljubeh, Marwan and Glaese, Mia and Jimenez, Carlos E. and Yang, John and
                  Ho, Leyton and Patwardhan, Tejal and Liu, Kevin and Madry, Aleksander},
  title        = {Introducing {SWE}-bench {V}erified},
  howpublished = {\url{https://openai.com/index/introducing-swe-bench-verified/}},
  year         = {2024},
  note         = {OpenAI blog post, August 13, 2024}
}

@misc{kimiteam2026kimik3openfrontier,
      title={Kimi K3: Open Frontier Intelligence}, 
      author={{Kimi Team}},
      year={2026},
      eprint={2607.24653},
      archivePrefix={arXiv},
      primaryClass={cs.CL},
      url={https://arxiv.org/abs/2607.24653}, 
}

\appendix
\renewcommand{\thetable}{\thesection\arabic{table}}
\renewcommand{\thefigure}{\thesection\arabic{figure}}
\counterwithin*{table}{section}
\counterwithin*{figure}{section}

\section{Experimental parameters and task selection}\label{app:A}

All runs use Kimi K3 through a Modal endpoint with an OpenAI-compatible interface, at temperature $1.0$, with \texttt{reasoning\_effort} set to \texttt{low}, \texttt{high} or \texttt{max}. The scaffold is mini-swe-agent \cite{mini_swe_agent}, executing each task inside the standard SWE-bench Verified Docker image for that instance. Repeats are independent samples: the endpoint accepts a seed parameter but does not return deterministic output for a fixed seed, so seeds index repetitions rather than reproducible draws.

Each run is bounded by a limit of $120$ agent turns and a ceiling of $4.00$ USD. We initially ran with a $60$-turn limit, which censored sparse specifications preferentially, since they take more turns and were therefore terminated more often. Every affected run was re-executed at $120$ turns, at which $108$ of the $112$ previously capped runs completed successfully.

\begin{table}[!ht]
\centering
\setlength{\tabcolsep}{4pt}
\small
\begin{tabular}{@{}lrrrr@{}}
\toprule
task & pooled solve rate & median cost & median turns & cost, low to max effort \\
\midrule
scikit-learn-14053 & $99.8\%$ & $0.111$ & 12 & $\times3.62$ \\
astropy-13579 & $99.6\%$ & $0.717$ & 30 & $\times4.35$ \\
xarray-7393 & $99.2\%$ & $0.461$ & 35 & $\times6.02$ \\
django-15503 & $86.3\%$ & $1.164$ & 38 & $\times4.19$ \\
astropy-14365 & $98.0\%$ & $0.181$ & 15 & $\times6.47$ \\
\bottomrule
\end{tabular}
\caption{The five tasks. Solve rate, cost and turns are pooled over the eleven analyzed specifications and three thinking efforts. Costs are USD at list prices.}
\label{tab:A1}
\end{table}

\section{The prompt set}\label{app:B}

Specifications follow the GitHub Spec Kit template \cite{speckit} and are stored as structured sections, from which all variants are emitted programmatically, so that a variant differs from the full specification only by the removal of whole sections. The eight sections are header, user story, acceptance scenarios, edge cases, functional requirements, key entities, success criteria, and assumptions. Figure~\ref{fig:1} shows two variants of one task; the following matrix gives the full design.

\begin{table}[!ht]
\centering
\setlength{\tabcolsep}{4pt}
\small
\begin{tabular}{@{}lr@{}}
\toprule
variant & sections retained \\
\midrule
full & all eight \\
no\_assumptions & all but assumptions \\
no\_entities & all but key entities \\
no\_edge & all but edge cases \\
no\_success & all but success criteria \\
no\_requirements & all but functional requirements \\
no\_acceptance & all but the acceptance scenarios, story prose retained \\
no\_userstory & all but the User Scenarios and Testing block, story and scenarios both \\
contract & header, functional requirements, success criteria \\
minimal & header, user story \\
raw & none; the original failing test output \\
oracle & none; the solution, with an instruction to apply it \\
\bottomrule
\end{tabular}
\caption{The twelve specifications, including two anchor prompts and ten structured variations comprising different specification sections.}
\label{tab:B1}
\end{table}

\begin{table}[!ht]
\centering
\setlength{\tabcolsep}{4pt}
\footnotesize
\resizebox{\linewidth}{!}{%
\begin{tabular}{@{}lrrrrr@{}}
\toprule
variant & astropy-13579 & astropy-14365 & django-15503 & scikit-learn-14053 & xarray-7393 \\
\midrule
full & 938 & 563 & 624 & 648 & 600 \\
no\_entities & 888 & 518 & 586 & 603 & 556 \\
no\_edge & 885 & 517 & 579 & 603 & 556 \\
no\_assumptions & 891 & 515 & 571 & 602 & 549 \\
no\_success & 854 & 501 & 561 & 595 & 546 \\
no\_requirements & 845 & 508 & 556 & 581 & 532 \\
no\_acceptance & 572 & 424 & 449 & 445 & 455 \\
no\_userstory & 372 & 294 & 310 & 298 & 305 \\
contract & 222 & 156 & 174 & 163 & 166 \\
minimal & 149 & 109 & 120 & 125 & 129 \\
raw & 573 & 957 & 957 & 983 & 483 \\
oracle & 295 & 273 & 865 & 173 & 185 \\
\bottomrule
\end{tabular}
}
\caption{Specification length in tokens, excluding a shared scaffold and system-prompt floor of approximately \emph{$1{,}100$} tokens present in every run. Counts are exact Kimi K3 tokens, derived from the server's first-turn input counts.}
\label{tab:B2}
\end{table}


\FloatBarrier
\begingroup
\begin{promptbox}{Header \textnormal{--- 88 tokens}}
\textbf{Feature Branch}: \texttt{fix/stack-preserves-coord-dtype}
\textbf{Created}: 2022-12-20
\textbf{Status}: Draft
\end{promptbox}
\vspace{2pt}

\begin{promptbox}{User story \textnormal{--- 111 tokens}}
\pheading{User Story 1 - Stack without changing coordinate dtypes (Priority: P1)}

As a user stacking dimensions into a MultiIndex, I need the coordinate dtypes to survive the operation, so that comparisons and downstream code that depend on a narrow integer type keep working.

\textbf{Why this priority}: The change is silent --- no error is raised, values simply come back with a wider dtype.

\textbf{Independent Test}: Build a dataset with an \texttt{int32} coordinate, stack it, and compare the coordinate dtype before and after.
\end{promptbox}
\vspace{2pt}

\begin{promptbox}{Acceptance scenarios \textnormal{--- 152 tokens}}
\textbf{Acceptance Scenarios}

\begin{itemize}[leftmargin=1.1em,itemsep=0pt,topsep=1pt,parsep=0pt]
\item \textbf{Given} a dataset whose \texttt{a} coordinate has dtype \texttt{i4}, \textbf{When} the dataset is stacked with \texttt{stack(b=('a',))}, \textbf{Then} the \texttt{a} coordinate still has dtype \texttt{i4}.
\end{itemize}

\begin{itemize}[leftmargin=1.1em,itemsep=0pt,topsep=1pt,parsep=0pt]
\item \textbf{Given} the same dataset, \textbf{When} the stacked result is unstacked, \textbf{Then} the coordinate dtype is still \texttt{i4}.
\end{itemize}
\begin{tcbverbatim}
   import xarray as xr
   import numpy as np

   ds = xr.Dataset(coords={'a': np.array([0], dtype='i4')})
   ds['a'].values.dtype == ds.stack(b=('a',))['a'].values.dtype   # expected: True
\end{tcbverbatim}
\end{promptbox}
\vspace{2pt}

\begin{promptbox}{Edge cases \textnormal{--- 27 tokens}}
\begin{itemize}[leftmargin=1.1em,itemsep=0pt,topsep=1pt,parsep=0pt]
\item Stacking several coordinates of differing dtypes must preserve each one independently.
\item Float and datetime coordinates must be unaffected.
\end{itemize}
\end{promptbox}
\vspace{2pt}

\begin{promptbox}{Functional requirements \textnormal{--- 52 tokens}}
\pheading{Functional Requirements}

\begin{itemize}[leftmargin=1.1em,itemsep=0pt,topsep=1pt,parsep=0pt]
\item \textbf{FR-001}: The system MUST preserve each coordinate's dtype when a MultiIndex is created by \texttt{stack}.
\item \textbf{FR-002}: The system MUST NOT alter values, only guarantee the dtype is carried through.
\end{itemize}
\end{promptbox}
\vspace{2pt}

\begin{promptbox}{Key entities \textnormal{--- 35 tokens}}
\begin{itemize}[leftmargin=1.1em,itemsep=0pt,topsep=1pt,parsep=0pt]
\item \textbf{stack} --- combines dimensions into a MultiIndex.
\item \textbf{MultiIndex level} --- the per-level array a stacked coordinate is read back from.
\end{itemize}
\end{promptbox}
\vspace{2pt}

\begin{promptbox}{Success criteria \textnormal{--- 48 tokens}}
\begin{itemize}[leftmargin=1.1em,itemsep=0pt,topsep=1pt,parsep=0pt]
\item \textbf{SC-001}: A coordinate created with dtype \texttt{i4} still reports dtype \texttt{i4} after the dataset is stacked.
\item \textbf{SC-002}: No existing stack or unstack behaviour regresses.
\end{itemize}
\end{promptbox}
\vspace{2pt}

\begin{promptbox}{Assumptions \textnormal{--- 57 tokens}}
\begin{itemize}[leftmargin=1.1em,itemsep=0pt,topsep=1pt,parsep=0pt]
\item Observed with xarray 2022.10.0, pandas 1.5.1, numpy 1.23.4.
\item pandas widens integer index types on MultiIndex construction; the fix is assumed to belong in xarray.
\end{itemize}
\end{promptbox}
\vspace{2pt}

\begin{promptbox}{Raw anchor \textnormal{--- the test output alone, 483 tokens}}
\begin{tcbverbatim}
============================= test session starts ==============================
platform linux -- Python 3.10.15, pytest-7.4.0, pluggy-1.5.0
rootdir: /testbed
configfile: setup.cfg
plugins: env-1.1.5, xdist-3.6.1, cov-5.0.0, timeout-2.3.1, hypothesis-6.115.5
collected 73 items

xarray/tests/test_indexes.py ........................................... [ 58%]
............................FF                                           [100%]

=================================== FAILURES ===================================
__________________ test_restore_dtype_on_multiindexes[int32] ___________________

        [ ... ]

        foo = xr.Dataset(coords={"bar": ("bar", np.array([0, 1], dtype=dtype))})
        foo = foo.stack(baz=("bar",))
>       assert str(foo["bar"].values.dtype) == dtype
E       AssertionError: assert 'float64' == 'float32'
E         - float32
E         + float64

/testbed/xarray/tests/test_indexes.py:706: AssertionError
=========================== short test summary info ============================
FAILED xarray/tests/test_indexes.py::test_restore_dtype_on_multiindexes[int32]
FAILED xarray/tests/test_indexes.py::test_restore_dtype_on_multiindexes[float32]
========================= 2 failed, 71 passed in 1.07s =========================
\end{tcbverbatim}
\end{promptbox}
\vspace{2pt}

\begin{promptbox}{Oracle anchor \textnormal{--- 185 tokens}}
\begin{tcbverbatim}
Make exactly this change to `xarray/core/indexing.py`, then submit. No investigation, reproduction, or verification is needed.

```diff
diff --git a/xarray/core/indexing.py b/xarray/core/indexing.py
--- a/xarray/core/indexing.py
+++ b/xarray/core/indexing.py
@@ -1531,8 +1531,12 @@ def __init__(
         self.level = level
 
     def __array__(self, dtype: DTypeLike = None) -> np.ndarray:
+        if dtype is None:
+            dtype = self.dtype
         if self.level is not None:
-            return self.array.get_level_values(self.level).values
+            return np.asarray(
+                self.array.get_level_values(self.level).values, dtype=dtype
+            )
         else:
             return super().__array__(dtype)
```
\end{tcbverbatim}
\end{promptbox}
\vspace{2pt}

\captionof{figure}{The full specification for \texttt{pydata/xarray-7393}, section by section, with both anchors. Each of the seven single-section variants removes exactly one of the first eight boxes; \texttt{contract} keeps the header, requirements and success criteria, and \texttt{minimal} keeps the header and user story, as set out in Table~\ref{tab:B1}. Token counts exclude the shared scaffold floor.}
\label{fig:B1}
\endgroup

Two features of this table qualify the reading of the anchors. The raw anchor is sparse in specification structure but is not short: on three of five tasks it is longer than the full specification, because a raw test transcript carries session headers, tracebacks and repeated assertions that a specification compresses. The oracle is short on four tasks but long on django-15503, where the fix itself is substantial. Neither anchors nor spec ablations are length-controlled.

Specifications were drafted with Fable 5 from the original task descriptions and then hand-edited for consistency and for faithfulness to the template.

\section{Statistical model}\label{app:D}

For each section and each outcome we fit

\begin{equation}
d_t \sim \mathrm{Normal}(\theta_t, s_t^2), \qquad \theta_t \sim \mathrm{Normal}(\theta, \sigma^2),
\end{equation}

where $d_t$ is the effect measured on task $t$, the difference in mean log cost or log turns against the full specification averaged over the three thinking efforts, and $s_t^2$ is its sampling variance. Priors are $\theta \sim \mathrm{Normal}(0, 0.5^2)$ and $\sigma \sim \mathrm{HalfNormal}(0.3)$. The task-level effects integrate out in closed form, so the posterior is evaluated on a $1201 \times 130$ grid over $(\theta, \sigma)$ rather than sampled. For solve rate, which is a proportion rather than a log quantity, both prior scales are halved and per-task proportions use the Agresti-Coull adjustment, without which a cell at $100\%$ contributes zero variance.

Both priors are weakly informative and stated in log units, so their scales are multiplicative. The prior on $\theta$ is centered at zero, with one standard deviation spanning a factor of $0.61$ to $1.65$; it expresses only that removing one section is unlikely to change cost by more than roughly a factor of $2.7$. The prior on $\sigma$ carries more weight, because with five tasks the between-task variance is barely identified and an unconstrained $\sigma$ drifts to implausible values, inflating every interval. Its median of $0.20$ corresponds to typical disagreement between tasks of around $22\%$.

\begin{table}[!ht]
\centering
\setlength{\tabcolsep}{4pt}
\small
\begin{tabular}{@{}lrrr@{}}
\toprule
prior & wholesale cut, cost & wholesale cut, turns & acceptance scenarios, turns \\
\midrule
$\tau=0.5$, $s=0.3$ (reported) & $+29.7\%$ $[+5.1, +58.7]$ & $+16.4\%$ $[+4.9, +30.7]$ & $+7.0\%$ $[+1.6, +14.1]$ \\
$\tau=1.0$, $s=0.6$ & $+31.0\%$ $[+3.3, +67.5]$ & $+16.6\%$ $[+4.1, +32.8]$ & $+7.0\%$ $[+1.4, +14.6]$ \\
$\tau=0.25$, $s=0.15$ & $+26.6\%$ $[+7.7, +46.8]$ & $+16.0\%$ $[+6.0, +27.1]$ & $+7.0\%$ $[+2.0, +13.2]$ \\
flat $\sigma$ (improper) & $+29.2\%$ $[+0.6, +63.2]$ & $+16.6\%$ $[+3.3, +32.3]$ & $+7.0\%$ $[+1.4, +14.6]$ \\
\bottomrule
\end{tabular}
\caption{Prior sensitivity. Every effect reported as established survives all four specifications. One effect we do not report as established, removing everything except requirements and success criteria, clears zero under the tightest prior alone at \emph{$+22.4\%$} \emph{$[+1.0, +46.5]$}; excluding it is therefore a conservative choice rather than a clean verdict.}
\label{tab:D1}
\end{table}

We report 90\% intervals rather than 95\%. At 95\%, two of the three effects we describe as established include zero: everything beyond the user story on cost runs from -0.3\% to +66.8\%, and the acceptance scenarios on turns from -0.1\% to +16.7\%. Because the verdict is sensitive to the threshold, we also report the posterior probability that each effect exceeds zero, which is 0.97 for cost and 0.98 for turns on everything beyond the user story, and 0.97 for the acceptance scenarios on turns.

Finally, we can contrast specifications against a fixed baseline but cannot rank them. Splitting the fifteen repeats of each cell in half and comparing the two orderings gives a mean Spearman correlation of $\rho = 0.41$, which the Spearman-Brown formula projects to $\rho = 0.58$ at the full fifteen. A stable ranking would need repeats an order of magnitude beyond what we ran.

\section{Per-section and per-task results}\label{app:E}

\begin{table}[!ht]
\centering
\setlength{\tabcolsep}{4pt}
\footnotesize
\resizebox{\linewidth}{!}{%
\begin{tabular}{@{}lrrrrr@{}}
\toprule
removing & $\Delta$ cost & $\Delta$ turns & $\Delta$ solve rate (pts) & task disagreement & tasks positive \\
\midrule
everything except the user story & $\bm{+29.7\%}$ $\bm{[+5.1, +58.7]}$ & $\bm{+16.4\%}$ $\bm{[+4.9, +30.7]}$ & $-1.9$ $[-7.4, +3.6]$ & $29\%$ & 5/5 \\
everything except requirements and success criteria & $+26.4\%$ $[-2.8, +62.6]$ & $+16.0\%$ $[-0.6, +35.3]$ & $-1.5$ $[-7.0, +4.0]$ & $40\%$ & 5/5 \\
acceptance scenarios & $+11.6\%$ $[-5.3, +32.3]$ & $\bm{+7.0\%}$ $\bm{[+1.6, +14.1]}$ & $-0.2$ $[-5.6, +5.1]$ & $22\%$ & 4/5 \\
user story and its scenarios & $+10.3\%$ $[-8.8, +34.2]$ & $+7.3\%$ $[-1.8, +19.0]$ & $+0.5$ $[-4.8, +5.8]$ & $26\%$ & 4/5 \\
functional requirements & $-2.4\%$ $[-11.8, +5.8]$ & $-0.6\%$ $[-5.3, +3.9]$ & $-0.1$ $[-5.5, +5.2]$ & $8\%$ & 2/5 \\
edge cases & $-2.6\%$ $[-9.5, +4.3]$ & $-1.4\%$ $[-5.8, +3.3]$ & $+0.5$ $[-4.8, +5.8]$ & $4\%$ & 1/5 \\
assumptions & $-3.1\%$ $[-9.9, +5.1]$ & $-0.4\%$ $[-5.1, +4.7]$ & $+1.0$ $[-4.3, +6.4]$ & $6\%$ & 1/5 \\
key entities & $-3.3\%$ $[-9.2, +2.8]$ & $-1.6\%$ $[-5.4, +2.2]$ & $+0.0$ $[-5.3, +5.3]$ & $4\%$ & 1/5 \\
success criteria & $-5.4\%$ $[-12.7, +2.4]$ & $-1.6\%$ $[-5.6, +2.4]$ & $-0.6$ $[-6.0, +4.8]$ & $7\%$ & 1/5 \\
\bottomrule
\end{tabular}
}
\caption{Effect of removing each part of the specification, with 90\% credible intervals. Bold marks an interval that excludes zero. Task disagreement is the posterior median of \emph{$\sigma$}, the spread of the effect between tasks, expressed on the same multiplicative scale as the cost column; the two largest cuts disagree between tasks by more than their own pooled effect, which is why we treat sensitivity as a property of the task. The final column counts tasks on which the cost effect is positive.}
\label{tab:E1}
\end{table}

Table~\ref{tab:E1} pools across tasks. Broken out by task, the two largest cuts are positive on all five. Reducing the specification to a bare user story raises cost by $13\%$ on scikit-learn-14053, $16\%$ on astropy-14365, $21\%$ on django-15503, $24\%$ on astropy-13579 and $115\%$ on xarray-7393. The acceptance-scenario effect is positive on four of the five, running from $-8\%$ on scikit-learn-14053 to $+53\%$ on xarray-7393. The five sections we cannot separate from noise move cost by no more than $18\%$ in absolute value on any task, and their signs differ between tasks, which is what the small between-task disagreement in Table~\ref{tab:E1} reflects.

\begin{figure}[!ht]
\centering
\includegraphics[width=1.0\linewidth]{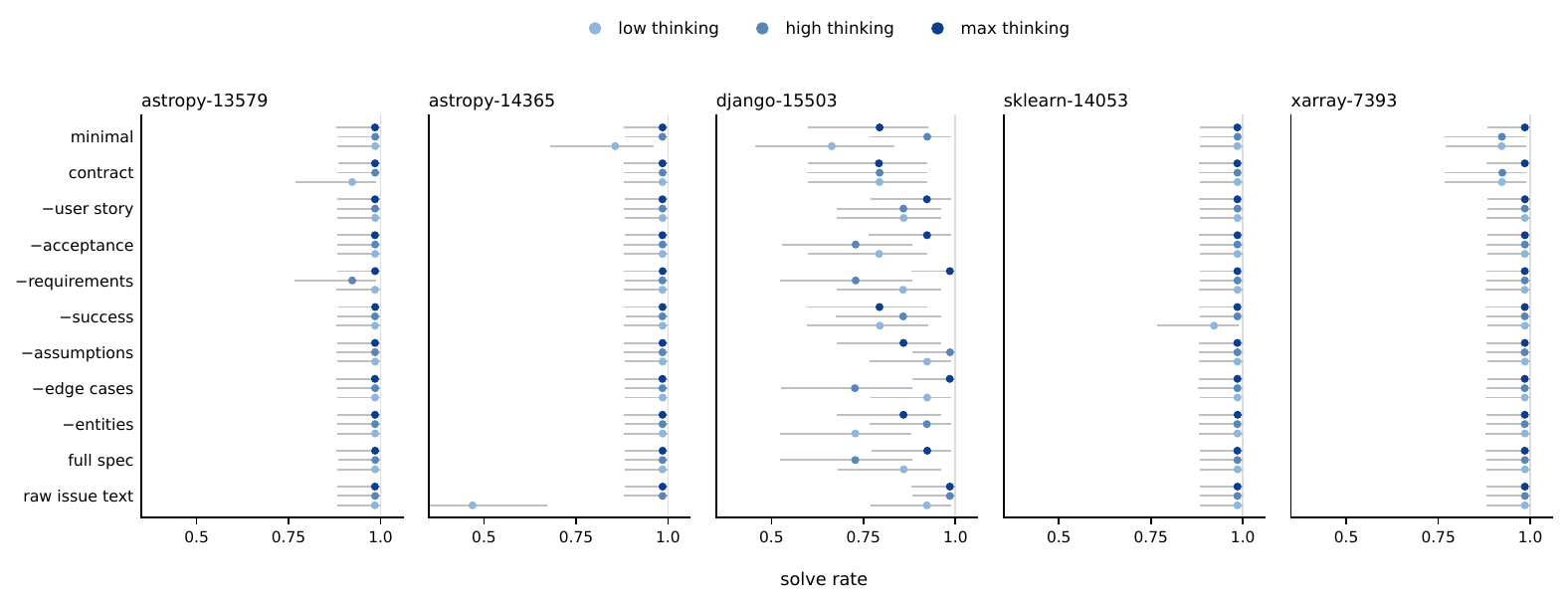}
\caption{Solve rate by specification and thinking effort for all five tasks, posterior median with $90\%$ credible intervals. Four of the five sit at or near the ceiling throughout. django-15503 has headroom under every specification and the raw issue text on astropy-14365 at low thinking effort resolves \emph{$47\%$} of the time against \emph{$99\%$} at high and max effort and \emph{$99.6\%$} pooled across the ten structured specifications on that task.}
\label{fig:F3}
\end{figure}

Per-task solve rates appear in Table~\ref{tab:A1} and, by specification and thinking effort, in Figure~\ref{fig:F3}. The same estimates expressed in tokens rather than dollars track the cost estimates closely, since log turns and log cost correlate at $r = 0.953$ over the solved runs.

\section{Token-denominated and supplementary figures}\label{app:F}

\begin{figure}[!ht]
\centering
\includegraphics[width=0.82\linewidth]{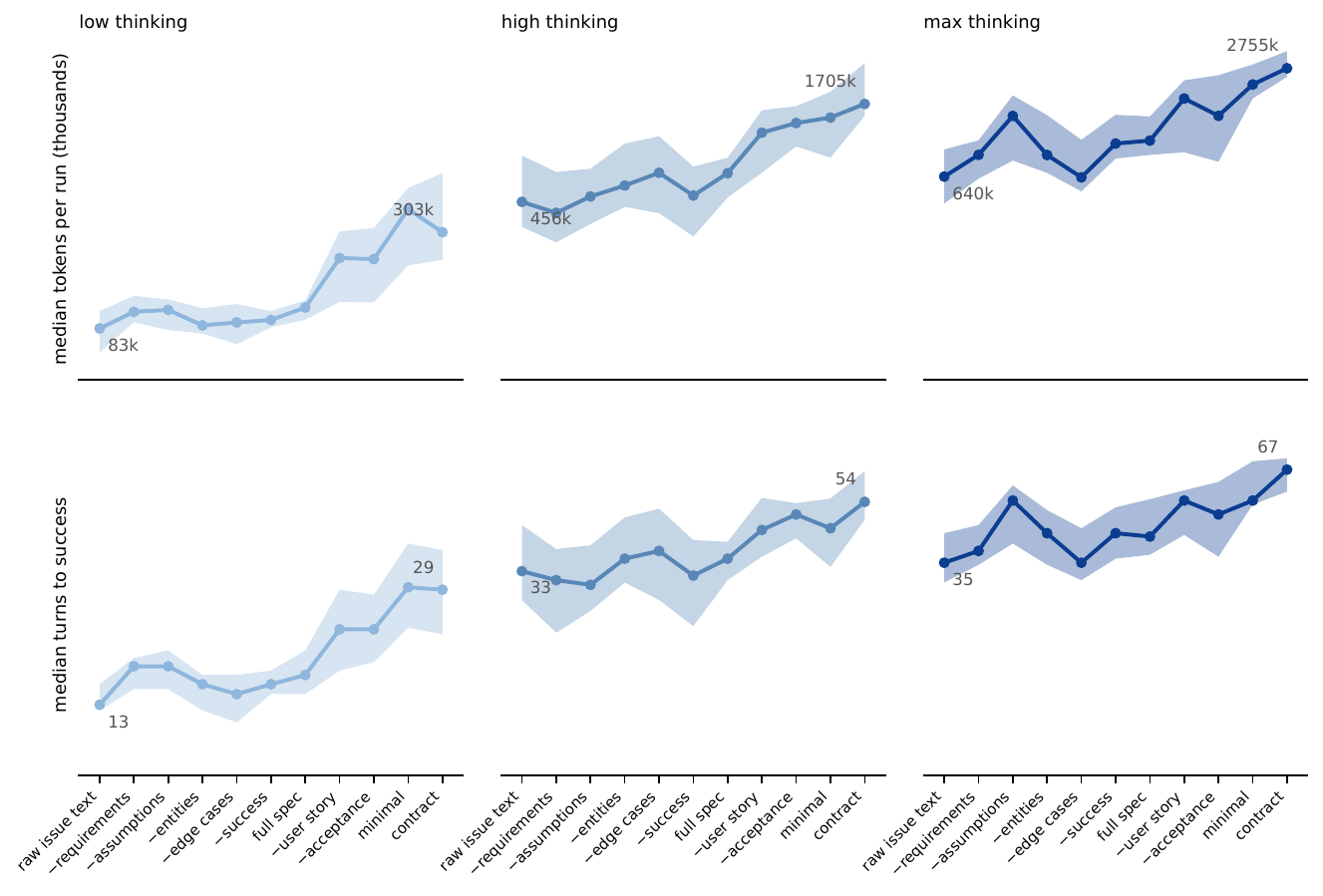}
\caption{Median tokens per run for xarray-7393, across the eleven specifications at each thinking effort.}
\label{fig:F1}
\end{figure}

Cost and tokens are near-equivalent under a single price schedule, but the mapping is provider-specific. Figure~\ref{fig:F1} repeats the cost figure, Figure~\ref{fig:2}, in tokens.

\begin{figure}[!ht]
\centering
\includegraphics[width=0.62\linewidth]{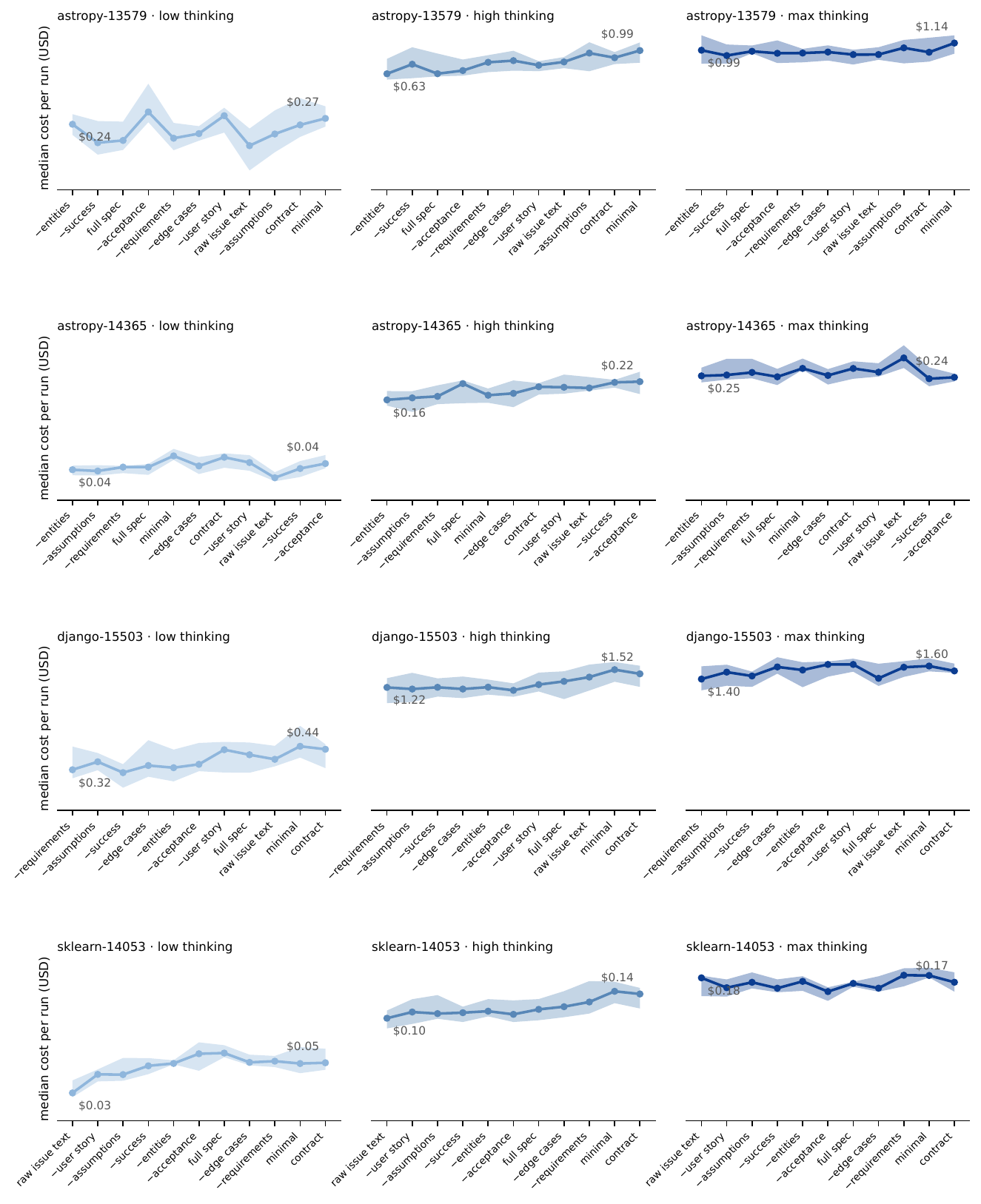}
\caption{Cost and turns for the four tasks not shown in Figure~\ref{fig:2}.}
\label{fig:F2}
\end{figure}

The composition of spend shifts slightly with thinking effort, as more reasoning buys more output tokens at the most expensive rate, but the shift is small relative to the effect of the price schedule.

\begin{table}[!ht]
\centering
\setlength{\tabcolsep}{4pt}
\small
\begin{tabular}{@{}lrrr@{}}
\toprule
 & fresh input & cached input & output \\
\midrule
tokens, low effort & $6.6\%$ & $89.9\%$ & $3.6\%$ \\
tokens, max effort & $3.1\%$ & $94.4\%$ & $2.5\%$ \\
dollars, low effort & $19.6\%$ & $26.8\%$ & $53.6\%$ \\
dollars, max effort & $12.3\%$ & $37.6\%$ & $50.0\%$ \\
\bottomrule
\end{tabular}
\caption{Composition of tokens and of spend, by thinking effort, at Kimi K3 list prices.}
\label{tab:F1}
\end{table}

\end{document}